\documentclass[11pt]{article}

\usepackage{float}
\usepackage{pslatex}
\usepackage{apacite}
\usepackage{amsfonts}       
\usepackage{amsmath}
\usepackage{graphicx}
\usepackage{setspace}
\usepackage{listings}
\usepackage{xcolor}

\title{'Memory States' from Almost Nothing:\\{\large Representing and Computing in a Non-associative Algebra}}

\author{
  St. Reimann$^{*}$\\
  \\
  Institute of Neuroinformatics, \\
  University of Zurich and ETH Zurich, \\
  Zurich, Switzerland
}

\begin{document}
% for the formatting instructions
\doublespacing

\maketitle

\begin{abstract} 

This note presents a non-associative algebraic framework for the representation and computation of information items in high - dimensional space. This framework is consistent with the principles of spatial computing and with the empirical findings in cognitive science about memory. Computations are performed through a process of multiplication-like binding and non-associative interference-like bundling. Models that rely on associative bundling typically lose order information, which necessitates the use of auxiliary order structures, such as position markers, to represent sequential information that is important for cognitive tasks. In contrast, the non-associative bundling proposed allows the construction of sparse representations of arbitrarily long sequences that maintain their temporal structure across arbitrary lengths. In this operation, noise is a constituent element of the representation of order information, rather than a means of obscuring it. The non-associative nature of the proposed framework results in the representation of a single sequence by two distinct states. The L-state, generated through left-associative bundling, continuously updates and emphasises a recency effect, while the R-state, formed through right-associative bundling, encodes finite sequences or chunks, capturing a primacy effect. The construction of these states may be associated with activity in the prefrontal cortex in relation to short-term memory and hippocampal encoding in long-term memory, respectively. The accuracy of retrieval is contingent upon a decision-making process that is based on the mutual information between the memory states and the cue. The model is able to replicate the Serial Position Curve, which reflects the empirical recency and primacy effects observed in cognitive experiments.\\


{\bf To appear} in {\it Neural Computation}, Vol 37, Issue 6, June 2025
\end{abstract}
    
\section{Introduction}
In essence, the perception of an object is initialised with the activation of a sensory pole. This sensory activation has a rapid decay and lasts for only a few milliseconds. Subsequently, selected parts are represented in neuronal structures, such as the prefrontal cortex, where elementary operations can be performed. 
Representations persist for a period of seconds. Ultimately, some of these representations are encoded and transferred into long-term memory, from which they can be retrieved and subsequently operationalised.
The objective of our model is not to provide a comprehensive account of these systems; rather, it offers a computational approach that aligns naturally with the tripartite memory process described. \\

In this process, representations are central, as they are induced by sensory activation on the one hand and encoded into memory on the other.
This note primarily concerns itself with the representation of information and the fundamental computations associated with it. Representing items as points in a high-dimensional space with an algebra defined it to compute such as in VSA-like structures in high-dimensional computing \cite{kanerva2009hyperdimensional} or in spatial computing \cite{lundqvist2023working} provide an elegant and straightforward framework for achieving this objective. 
The general framework is as follows: The perception of a physical item evokes a neural activity pattern in the brain, which is represented by a high-dimensional binary random variable. The realisations of this random variable are called states. These can be considered distributed representations of the perceived object. The state space is given a metric to quantify similarity. Computations are performed using addition for bundling and multiplication for binding items of information together. For an overview of the various HD algebras and their comparisons, please refer to \cite{schlegel2020comparison}.

While it is essential for many cognitive tasks to represent the sequential order in which items appear faithfully, sequential information is lost in the course of bundling if this operation is associative. This problem might be 'solved' by introducing additional structure, e.g. by postulating a set of position markers, which is sufficiently rich and ordered. The sequence $(a,b,c)$ is finally represented by binding sequence items and markers together $( a*1, b*2, c*3)$. To represent sequential order in relative terms, the sequence might be encoded using chaining: The representation becomes $(a*b, b*c, c)$ so that $a$ points to $b$ points to $c$. An other 'solution' relies on the postulate of activity or attention gradients \cite{page1998primacy}, which vary monotonically over time, e.g. ${\cal A}(x) < {\cal A}(y) < {\cal A}(z)$ indicates that $z$ is more recent than $y$, which is more recent than $x$. 
Similarly, in the TCM model \cite{sederberg2008context} the primacy gradient is assumed to be due to an exponential decay in the learning rate for early serial items. 
%s
The model for "storage and retrieval of items and associative information" proposed by B. Murdock in 1982 \cite{murdock1982theory,lewandowsky1989memory} is a full-blown VSA-model in today's terms. 
Since in his model, bundling is vector-addition and therefore associative, it is necessary to explicitly code serial order by exponentially varying weights over serial position.
In traditional VSA algebras, the additional structure erected is a permutation operator $\rho$, so that sequential order is represented by increasing powers of $\rho$, i.e. $(a,b,c)$ is represented by bundling $(\rho^2a, \rho b, c)$. For a detailed discussion see \cite{gallant2013representing}.
The subsequent unbinding, e.g. for recall, is performed by applying increasing powers of the inverse $\rho^{-1}$. For retrieval, $\rho$'s together with their respective inverses must be stored or computed. Although this is fairly easy to achieve in technical systems, it might be difficult for biological systems. 

Without question, associativity as resulting from bundling by vector-addition is computationally very convenient. But the price to be paid is the loss of order information and the effort of additional algebraic structure to re-establish serial information. So, if associativity of bundling creates this problem, overcome this model assumption.
This is the central theme in this note: Consider bundling as an interference-like non-associative operation and study the emerging properties of the algebra. 
\begin{equation}
    \Big(\mathbb{X}, *, +_\theta, \Big), 
\end{equation}
where $\mathbb{X}_q$ is a space of states, $*$ is multiplication-like binding, while $+_\theta$ is interference - like bundling. The mean activity of a state $x$ is denoted by $Q(x)$ and might serve as a measure for its sparseness. States having mean activity $q$ are called q-states.

Being high-dimensional, sparse states are biologically plausible \cite{olshausen2004sparse} as well as computationally efficient. Sparse representations enhance representational capacity allowing for efficient computations, while energy consumption \cite{howarth2012updated} is minimised, as evidenced by the findings of Foldiak \cite{foldiak2003sparse} and Palm \cite{palm2013neural}. A question is about sparseness can be locally controlled, therefore opening the opportunity for fine-tuning. I will show that in his model the bundling operation proposed allows to control sparseness.

By non-associative bundling sequential information is preserved through the process of bundling, so that the resulting state represents information about the sequence. Non-associative addition can be both, left and right associative. Without having a clear preference for one or the other, we assume that if both 'formal twins' are possible, both will be realised. Consequently, a unique sequence is represented by two different bundles: the $\bf L$-state results from left-associative addition, while the $\bf R$-state results from right-associative addition. Both states differ concerning represented information as well as from the viewpoint of construction.
As both states are related to the same sequence of items, they might be regarded as two components of the corresponding memory state
\begin{equation}\label{eq:M}
{\bf M} =\begin{pmatrix} {\bf L }\\ {\bf R} \end{pmatrix}.
\end{equation}
Thus, the non-associativity of bundling enriches the representation of sequential information by leveraging these dual associative processes, rather than requiring additional operations.  

The process of information retrieval might highlight the meaning of the memory state $\bf M$. In the framework outlined, this is due to a decision process. The decision is based on the comparison of some information item or cue against recent information, which is represented by the memory state $\bf M$.
This vector can be said to represent recent information. When the agent is presented a new information item or cue item, it evaluates the degree to which that item is already represented within the memory state.
This evaluation is quantified using mutual information between recent and new information. If the cue is new, i.e. dissimilar to information items already represented in $\bf M$, mutual information is zero. If it has been seen recently, it will be well represented in the $\bf M$ state and mutual information is high. The accuracy of identifying the cue as part of the sequence improves with the quality of its representation. The higher mutual information, the more likely it is that the agent determines the cue as having seen before. Note that accordingly new items that are sufficiently similar to old ones might be identified as known, while those even identical to those presented much earlier might not be correctly identified. Such recognition failures are thus natural consequences of the framework proposed. 

However, this mechanism naturally produces the characteristic Serial Position Curve (SPC), which describes the probability of correct recall based on an item's position within a sequence. Our model captures the typically observed asymmetric $\cup$-shaped SPC as a GenericBehaviour, where items from both the beginning (primacy effect) and end (recency effect) of the sequence are more accurately recalled than those in the middle, as shown in Fig. \ref{fig:GenericBehaviour}.

\section{An algebra on representations}
\subsection{States are sparse high-dimensional distributed representations of objects.}
The sensory activation pattern that is induced by the perception of an object, for example in the retina, is mapped on a receptive field in a region of the brain. Typically, the number of neurons in the receptive field is significantly much larger than the number of sensory cells. For the sake of simplicity, this receptive field is modelled as a grid of neurons, each of which is either active or inactive. A state $x$ is a configuration of the grid, and may be regarded as a realisation of a binary random variable $X$ defined on $G$. It is assumed that the grid size is sufficiently large. The object, designated by the variable $\bf x$, is then represented by a high-dimensional binary state $x$. This state provides a distributed, high-dimensional representation of the object in question within the population of neurons of the receptive field \cite{bays2024representation}. 

Given that representations are high-dimensional, two arbitrarily chosen states will be quasi-orthogonal, i.e. independent. Conversely, similarity should be represented accurately, whereby similar objects should be represented by near-by states in close proximity, see constraint $4$ in \cite{gallant2013representing}. The representation is assumed to be continuous. 

In accordance with the concept of spatial computation as outlined in reference \cite{lundqvist2023working}, information pertaining to items is 'spatially organised' within this metric state-space, whose geometry is according to some distance or similarity. Computations involving these items, such as bundling and binding, can be regarded as 'movements' within that space, see Fig \ref{fig:movement} and the sections below.  

\subsection{Binding and bundling are elementary operations on representations} 
A state is a holographic representation of an object, e.g. a face. This object might be linked to a particular context and might be associated with a particular feature. This feature again is represented by a state. The object linked to some context should also be a state. Two associated items $x$ and $y$ are represented by the state $z$ which results from binding $x$ to $y$, i.e. $z = x*y$. Being concerned with binary states, a particularly simple {\bf binding} operation is the component-wise XNOR: Two states are more tightly bond together the more they coincide in $0$'s and $1$'s, i.e. in the simultaneously active or inactive sites. In this sense, binding is by coincidence detection. Under this operation, every state is self-inverse, i.e.
\begin{equation}\label{eq:si}
    x*x=1,
\end{equation}, where $1$ is the one-vector. This property will become important later for dissolving a bound state into single items, $x*(y*x)=y$. Furthermore, it does not come as a surprise that binding and the normed Hamming distance are closely related. In fact it holds
\begin{equation}\label{eq:distQ}
    d(x,y) = 1 - Q(x*y)
\end{equation}
The mean activity of the bound state serves as a measure for the dissimilarity of the respective states. 
For two independent random q-states $x,y$, their distance yields
\begin{eqnarray}
    d(x,y) &=& 2 q(1-q) \\
    d(x,x*y) &\stackrel{(4)}{=}& 1 - Q(x*x*y) \;\stackrel{(3)}{=}\; 1-Q(1*y) \nonumber \\
    &=& 1 - q
\end{eqnarray}

\begin{figure}
    \centering
    \includegraphics[width=0.6\linewidth]{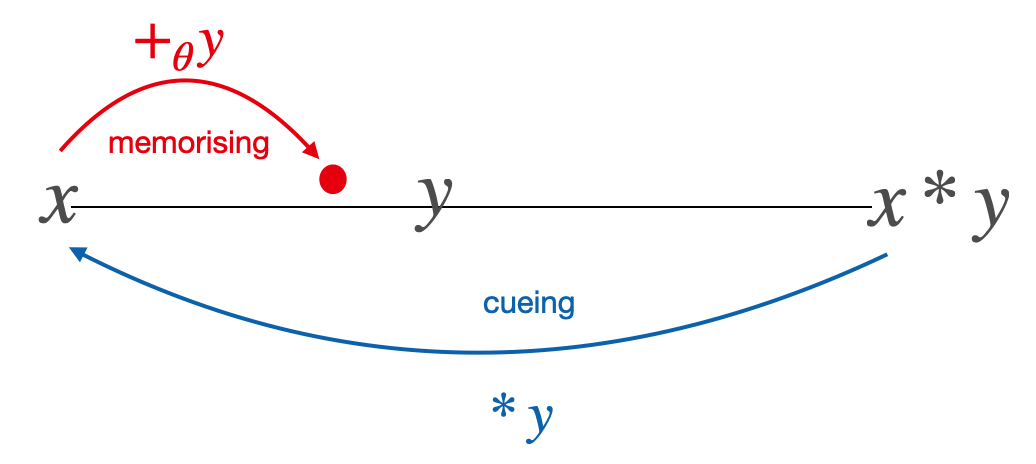}
    \caption{Computation by bundling or binding can be geometrically regarded as movement in state-space.}
    \label{fig:movement}
\end{figure}

The second operation on the state space is {\bf bundling} $z = x +_\theta y$. Bundling can be seen as a noisy interference of the states $x$ and $y$. It is component-wise defined according to: $0+_\theta 0 := 1$, $1+_\theta1 :=1$, while $0+_\theta 1= 1+_\theta 0 := 1$ with probability $\theta$. Note that $x+_\theta x=x$ for any state. However, bundling therefore introduces some noise into the calculation. 
This operation is non-associative unless $\theta=0,1$. For $\theta=0$, the bundling operation is a component-wise OR, while for $\theta = 1$, bundling equals component-wise AND. In these cases the operation is of course associative. 
This bundling operation generalises the one used in Binary Sparse Distributed Codes \cite{rachkovskij2001representation}. It should be noted that in the Binary Spatter Code \cite{kanerva1994spatter}, addition is implemented by a deterministic normalisation procedure.
It holds 
\begin{equation}
    d(x,x+_\theta y) = 2 q (1-q) (1-\theta)
\end{equation} 
for q-states $x,y$ and bundling parameter $0 \leq \theta \leq 1$. 
Finally in high dimensions, binding distributes over bundling. For q-states $x,y,z$, the states $a=x*(y+_\theta z)$ and $b=x*y +_\theta  x*z$ have distance $d(x,y) = q(1-q) \leq \frac{1}{4}$, while the probability that two independently chosen states in $N$ dimensions have a distance smaller than $\frac{1}{4}$ decreases exponentially to zero when $N$ increases. So even for moderate dimensions, $P(d \leq \frac{1}{4}) \approx 0$. For large $N$ we therefore write
\begin{equation}\label{eq:distributivity}
    x*(y+z) \doteq x*y + x*y. 
\end{equation}
The following is often paraphrased as "Addition creates similarity, while multiplication creates dis-similarity."
\begin{equation}
    d(x+_\theta y,x) \leq d(x,y) \leq d(x*y,x) \qquad 0 < q \leq \frac{1}{2}
\end{equation}
From a geometrical point of view these computations can be seen as generating translations in state-space. Given state $x$, bundling to it state $y$ corresponds to a movement 'towards' $y$, while binding $y$ to the bond state $x*y$ throws it to $x$, see Fig. \ref{fig:movement}. This relation between computation and 'movement' corresponds to the main assumption in spatial computing that, depending on the particular task to be achieved, WM item representations [states] are consistently moved in 'memory space' [state space] \cite {lundqvist2023working}.

\subsection{Mutual information between states $x$ and $y$ is related to the mean activity of their bond state $x*y$.}
We will employ mutual information in our model to quantify the degree to which one state represents another. The higher the mutual information $I(x;y)$ between states $x$ and $y$, the better $x$ represents $y$ and vice versa. While our aim is to build our model exclusively on the elementary operations in our algebra, the question is how to calculate or at least approximate mutual information by these elementary operations? 
It can be shown analytically that in our high dimensional setting mutual information between two binary states $x$ and $y$ can be well approximated by a quadratic function of the mean activity of the bond state $x*y$, i.e.
\begin{equation}\label{eq:MI_Q}
   I(x;y) = 8q(1-q)  \left(\frac{1}{2} - Q(x*y)\right)^{2} + O\big(Q(x*y)^4\big), 
\end{equation}

The argument is as follows. Let the state $y$ be derived from a q-state $x$ by flipping a portion $\epsilon$ of its bits. In this case $d(x,y) = \epsilon$, while its mean activity yields $Q(y) = \epsilon q$. 
Their joint probability distribution is
\[
\Phi(x,y) = \begin{pmatrix}
    (1-q)(1-\epsilon) & (1-q)\epsilon\\
    q\epsilon & q (1-\epsilon)
\end{pmatrix}
\]
Since $\det \Phi(x,y) = 2q(1-q)(\frac{1}{2}-\epsilon)$ and $\epsilon = d(x,y)$, a straight forward calculation reveals that mutual information is  approximated in first order by 
\begin{equation}\label{eq:MIDist}
   I(x;y) = 8q(1-q) \left( d(x,y) - \frac{1}{2}\right)^{2} + \hdots .
\end{equation}
Mutual information is maximal for $d(x,y) = \{1,0\}$, i.e. states are perfectly correlated or anti-correlated, respectively and minimal if states are quasi-orthogonal $d(x,y)=\frac{1}{2}$, i.e. half of the components in $x$ are flipped. Near-by-states, therefore, have high mutual information. This agrees with the intuition that if the state $y$ is close to $x$, knowing $x$ already tells a lot about $y$. Using eq. \ref{eq:distQ} yields eq. \ref{eq:MI_Q}.
\begin{figure}
    \centering
    \includegraphics[height=2in]{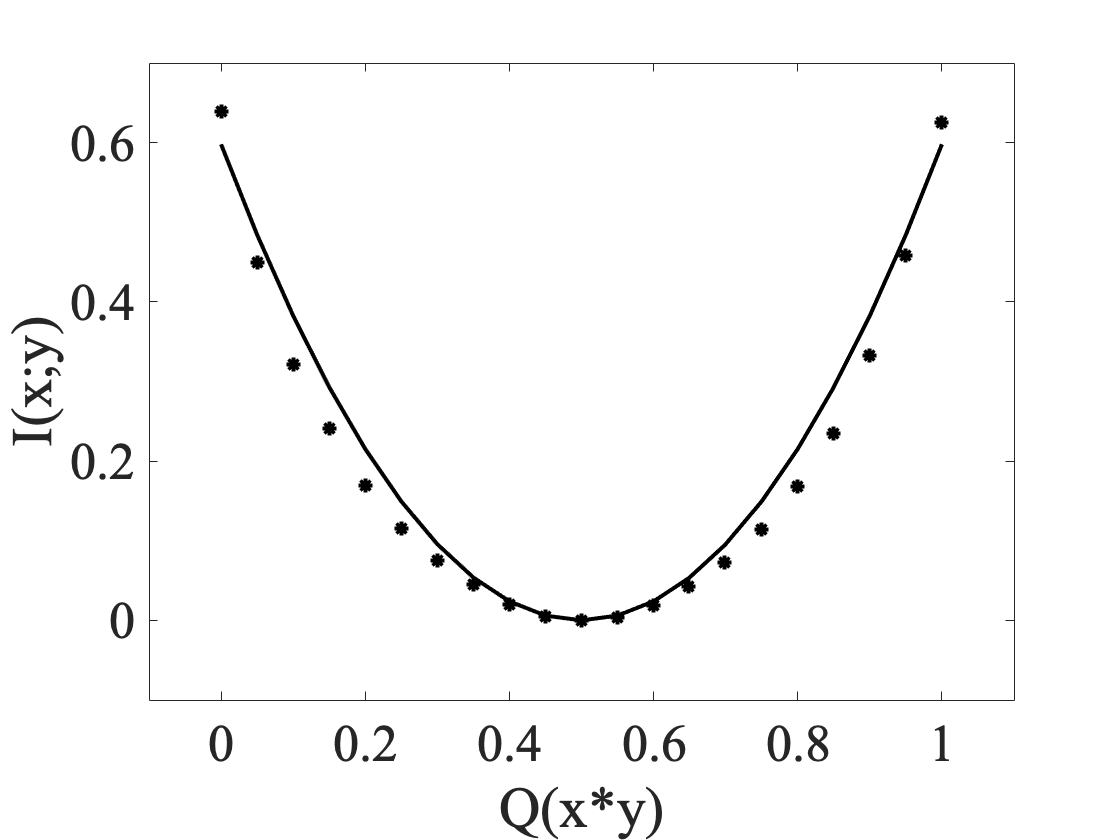}
    \caption{ The mutual information $\bullet$ between states varying with their distance $\epsilon$ and its approximation by the mean activity of the bound state (solid line) according to eq. \ref{eq:MI_Q}.
    }
    \label{fig:IQ}
\end{figure}

\section{Representing sequences}
\subsection{The internal representation of a finite sequence}

In experiments, finite sequences $(A,B,\hdots,E)$ are presented to an agent.
The perceived elements are represented by states $\{a,b, \hdots,e\}$. Due to left and right associative bundling, the sequence is represented by two states \cite{reimann2021algebra} 
\begin{eqnarray}
	\bf {L} &=& \big(\hdots(\eta + a) + b ) + c \big)+ d ) + e \label{eq:L-state}\\
	\bf{R} &=& \eta + \big(a + ( b +  ( c + ( d + e)\hdots\big) \label{eq:R-state}.
\end{eqnarray}
where $\eta$ is the initial state of the medium in which the sequence is represented. 
These states become orthogonal if the length of the memory list increases. The reason is that the longer the sequence is, the more uncorrelated noise is accumulated by bundling, so that the two sequences become increasingly uncorrelated to each other.

\begin{figure}
    \centering
    \includegraphics[width=4.3in]{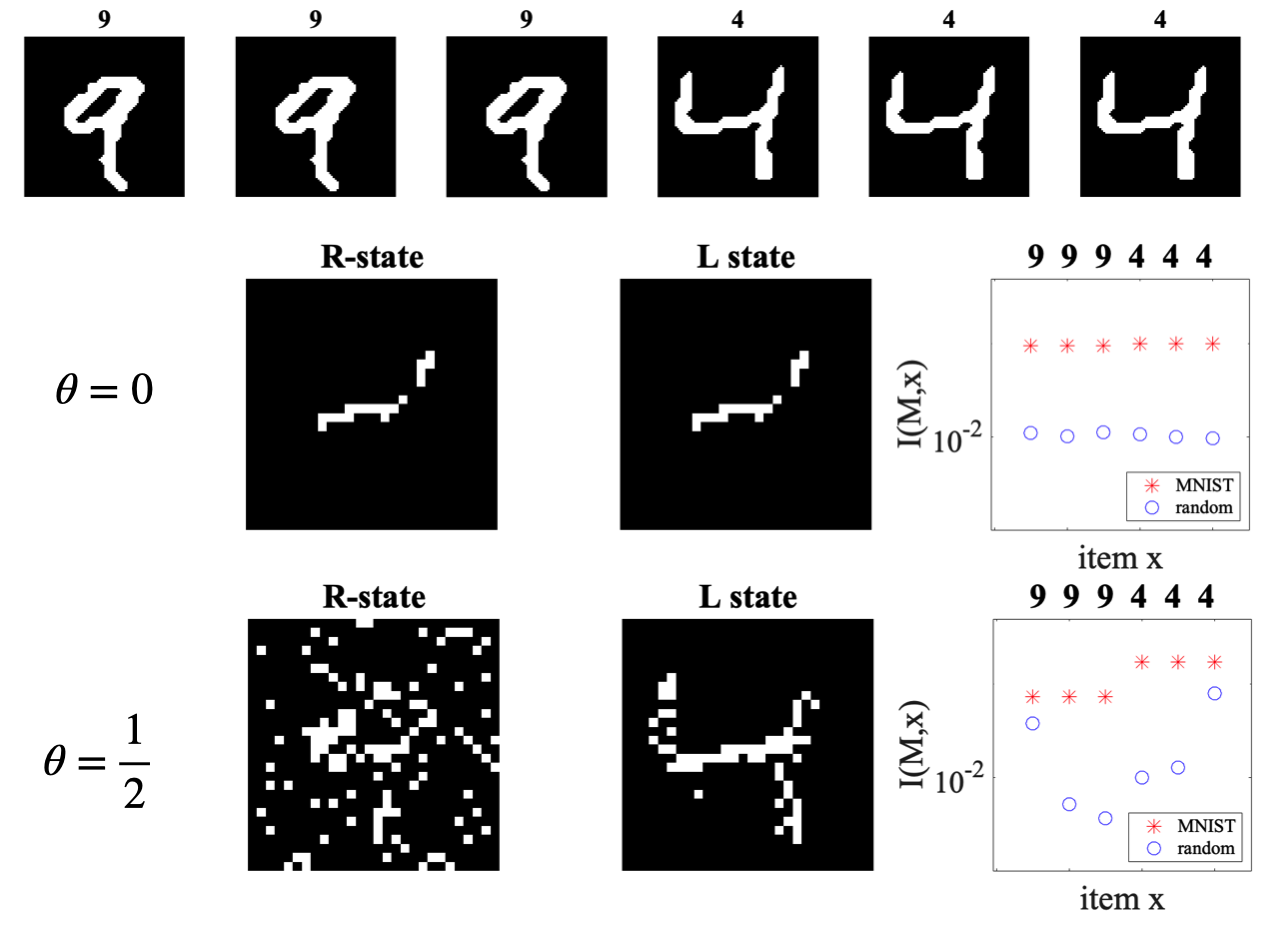}
    \caption{ The states $\bf R$ and $\bf L$ have grid size $28 \times 28$ (original MNIST figures) and result from bundling the items in the sequence using $+_\theta$ for $\theta=1/2$ (non-associative) and $\theta=0$ (associative), respectively. Associative bundling does not reveal sequential order information, while non-associative bundling does. $I({\bf M},x)$ is the mutual information between the state $M$ and some item $x$, see eq. \ref{eq:SPC}. The profiles for MNIST states (red *) and random IID states (blue $\circ$ are shown in the most right column.}
    \label{fig:999444}
\end{figure}

The two states are very different in nature: While the $\bf L$ state can be up-dated iteratively on the fly, the $\bf R$ state cannot be constructed unless the entire sequence is presented. Therefore, an ongoing inflow of information can be represented by $\bf L$, while in the $\bf R$-state only finite chunks of information can be represented. This is fulfilled in experiments where only finite sequences are presented. Chunking is one of the methods to rehearsal the memory of a text is chunking, see \cite{gobet2001chunking}. The entire text is broken down into smaller sections or chunks, where then the focus is on memorising one chunk at a time before moving on to the next. These two states are the components of the state $\bf M$ representing the sequence
\begin{equation}
    \bf{M} = \begin{pmatrix}
        \bf{L}\\ \bf{R}
    \end{pmatrix}
\end{equation} 
In Fig \ref{fig:999444} I have used a sequence of original MNIST pictures of numbers to illustrate how the resulting $\bf L$ and the $\bf R$ states look  and, particularly, that they represent the sequence in a some-what inverse order. Obviously, if the operation is associative, both states are identical. In the case of non-associativity, the most recent item $4$ is best represented by the $\bf L$ state, while the oldest item $9$ is best represented by the $\bf R$ state.
For the sake of readability, we may use the following convention. We write the 'state' representing the sequence by
\begin{equation}\label{eq:Convention}
X=\eta + a + b  + c + d + e,    
\end{equation}
which when specifying $\bf L$ or $\bf R$ becomes the respective state by correspondingly inserting the brackets.

\subsection{Sparseness of representations is controlled by bundling }
In artificial settings, such as experiments, the list length is finite or even restricted to very low numbers such as $5$ or $8$. In real world, the cognitive agent is subject to a continuous inflow of information, i.e. the input sequence has arbitrary length. 
What happens to a state if more and more items are bundled? If it would become more dense and more energy would be required. If it becomes homogeneous, it loses all the item information previously represented. Both do not happen if the bundling operation is governed by a threshold satisfying $\frac{1}{2} \leq \theta < 1$. In this case, an arbitrarily long sequence of q-states can be bundled together so that its serial order is preserved; the sparsity of that state increases, while the respective state does not become homogeneous $0$.

The main result is the following: If $\frac{1}{2} < \theta < 1$, arbitrarily long sequences have sparse, faithful representations with mean activity $q \leq \overline{Q} < 1$ (see below). 
If the list length increases, the mean activity of the memory state exponentially converges to an asymptotic value $\overline{Q}$ exponentially
\begin{equation}\label{eq:Q}
\overline{Q} := \frac{\theta \: q}{1-\theta(1-q)-q(1-\theta)}.
\end{equation}
Particularly, if $\frac{1}{2} \leq \theta < 1$, the asymptotic mean activity is bounded by $0 < \overline{Q} \leq q$.

\begin{figure}
{\center
\includegraphics[height=2.2in]{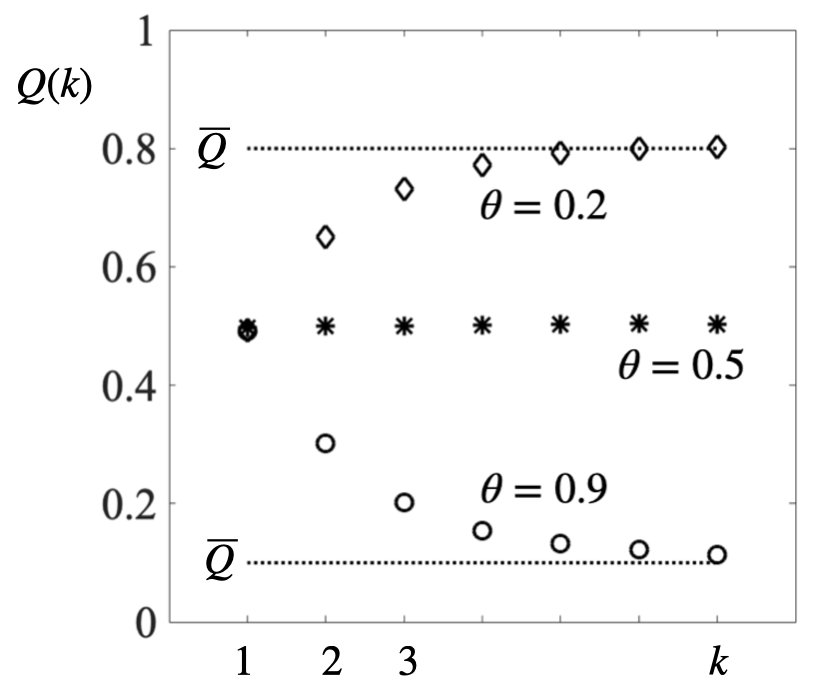} 
\caption{ {\bf Bundles remain sparse.} Mean activity of the state resulting from bundling $k$ dense quasi-orthogonal items is a monotonous function of list length due to $Q(k) = \theta - \left( \theta-\frac{1}{2}\right) 2^{1-k}$. Note that the degree of sparseness is controlled by the bundling parameter $\theta$. Dashed lines are the respective asymptotic values according to eq. \ref{eq:Q}. \label{fig:SparseMemory}}
}
\end{figure}

To guarantee, that states representing sequences of arbitrary length are sparse, we restrict the parameter range to
\begin{equation}
    0 < q \leq \frac{1}{2} \qquad \frac{1}{2} \leq \theta < 1
\end{equation} 
\subsection{How to retrieve item-, context-, as well as serial order information}
Having established the methodology for bundling information items to represent a sequence, we now turn our attention to the question of how that information can be retrieved from its representation. In the standard VSA setting, this task is solved by an iterative process of unbinding information items from a storage medium. In our approach, information from the past functions as a filter for information in the present.
\begin{figure}
{\center 
\includegraphics[width = 0.9 \linewidth]{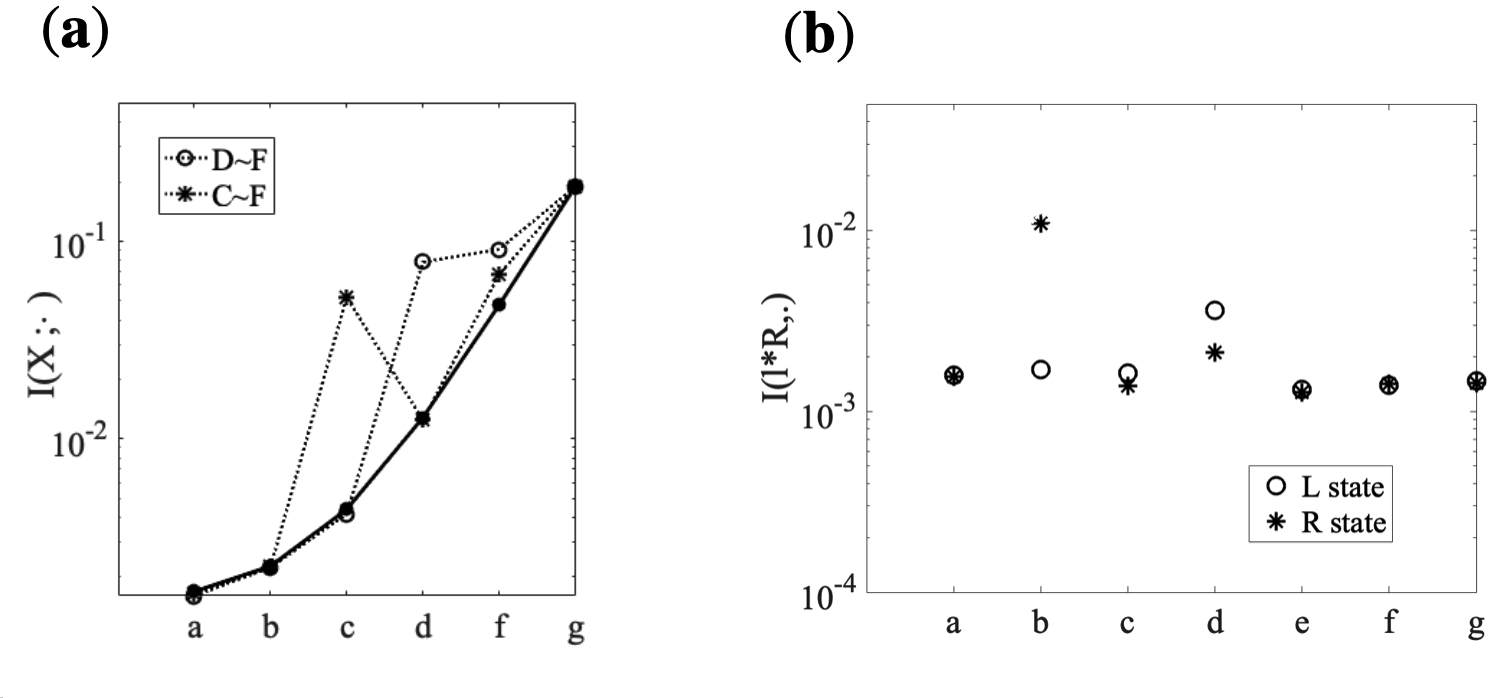}
\caption{ $\bf(a)$: Mutual information between states and the $\bf L$-state due to eq. \ref{eq:L-state} is an increasing function of their sequential order (solid line). Mutual information of similar items is similar, see $\circ$ for the case that $D \sim F$ and $\ast$ for the case that $C \sim F$. 
$\bf (b)$ Given $X=\eta + a*k + b*\ell  + c*m + d*\ell + e*n$, the cue $\ell$ has two values, $b$ and $d$. The more recent context $D$ is better represented in the $\bf L$- state due to the recency gradient. }\label{fig:MI-sequence}
}
\end{figure}
Let us consider the case of an agent that has constructed an internal representation, which we shall denote as $X$, of a set of information items arranged in a specific order. As the agent encounters various items, the key question is to ascertain which of these items are already known to the agent, that is, which ones are already well-represented by $X$. In order to quantify the degree to which an item, designated as $x$, is represented by the internal representation denoted as $X$, the mutual information $I(X,x)$ is employed. Various items are then ordered according to their respective mutual information (in ascending order) with respect to the internal representation, resulting in a sequence of items that are most similar to those in the list represented by $X$. This is illustrated in Fig.\ref{fig:MI-sequence} $\bf (a)$. It can thus be seen that an internal representation $X$ may be regarded as a filter which assigns positive values to items that are well represented, that is to say, items that are aligned with the preceding information. In this approach, the role of $X$ is not that of a storage medium from which content can be retrieved. Rather, it functions as a filter that compares new information items with previously constructed knowledge. In this manner, both item information and sequential information are retrieved. 

Context information is retrieved analogously after cueing. Cueing uncovers the context or context of that item, which is similar or even identical to the cue item. Mutual information between the cued state and other items reveals the one, which occurs in that context. The context may include various information modalities including spatial, temporal location or neighborhoods. As examples, two mechanisms to represent a sequence either by labeling items by serial position markers or by chaining are considered. Their respective representations being 
 \begin{eqnarray}
    Y &=& \eta + a*1 + b*2  + c*3 + \hdots\\
    Z &=& \eta + a*\eta + b*a  + c*b + \hdots
\end{eqnarray}
where the convention eq. \ref{eq:Convention} was used to simplify notation. Cueing the states $X,Y$ by $b$ respectely reveals
\begin{eqnarray}
    b*Y &\doteq& b*\eta + b*a + 2 + b*c+ b*\hdots\\
    b*Z &\doteq& b*\eta + b*a + a + c + b*c+\hdots
\end{eqnarray}
The action of cueing with an item thus is to uncover its respective context.
$I(\cdot, b*Y)$ has a peak at $2$, which shows that $B$ occurs in context (position) $2$. $I(\cdot, b*Z)$ has peaks in $a$ and $c$, saying that $B$ has two sequential neighbors, $A$ and $C$. Particularly, if $X=L$, then $I(a,b*Z) > I(c,b*Z)$ saying that $a$ is more recent than $c$.
Analogous to before, given a cue $c$, $c*X$ serves as a kind of filter to single out those items which occur as contexts in the internal representation. 

\subsection{The Serial Position Curve or: How much of an item is represented by the memory state?}\label{sec:SPC}
\begin{figure}
\includegraphics[width = 0.9 \linewidth]{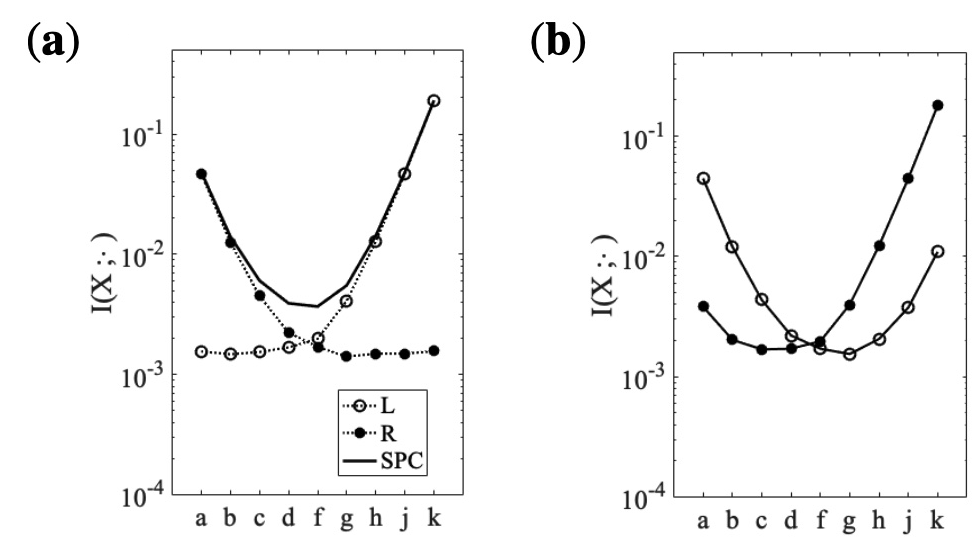}
\caption{
{\bf Simulation of Serial Position Curves}: {Mutual information $I(X;x)$ is used as a proxy for the accuracy of recalling item $x$ from the memory state. The SPC's are in good qualitative agreement with experimental data for free recall and seriual recall under forward or backward condition. }
}
\label{fig:GenericBehaviour}
\end{figure}

The present study applies the aforementioned model to derive the asymmetric and concave-up shape of the Serial Position Curve (SPC). The SPC demonstrates the accuracy of item retrieval as a function of serial position in a memory list, averaged over a sample of participants. The SPC exhibits a concave-up shape and is asymmetric \cite{murdock1974human, kahana2012foundations}. It is one of the benchmark findings in research on human working memory \cite{oberauer2018benchmarks} and was also found in rats \cite{kesner1982serial,Bolhuis1988serial}. It has been observed for several sensory modalities such as auditory and visual presentation, lip-reading, and sign language across different cultures. This phenomenon is robust to a wide range of actions such as drug intake, and is resistant to head injury, amnesia, Parkinson's disease, see \cite{capitani1992recency}. Its robustness and ubiquity speak for a very fundamental realisation by elementary mechanism without the need for elaborated and therefore vulnerable structures.
Experimental conditions and also other factors may affect the SPC. In free recall, after the presentation of a list of items, subjects were allowed to recall as many items as possible in any order. The SPCs for lists with lengths $10$ and $15$ \cite{murdock1974human} are shown in Fig. \ref{fig:GenericBehaviour} $\bf (a)$ as blue and black curves, respectively. SPC's in forward and backward paradigms \cite{oberauer2018benchmarks} differ in the manifestations of the primacy and recency effects, respectively, see Fig. \ref{fig:GenericBehaviour} {\bf (b)}. 

A single list is faithfully represented by the components of a state ${\bf M }= ({\bf L},{\bf R})$. Both states make a contribution to the overall mutual information relative to some cue $x$. 
The serial position curves results from a convex linear combination of the primacy gradient due to the $\bf R$-state and the recency gradient due to $\bf L$ according to.
\begin{equation}\label{eq:SPC}
I({\bf M};x) =  \rho \: I({\bf R};x) + \rho' \: I({\bf L};x)
\end{equation}
for a weight parameters $0 \leq \rho, \rho' \leq 1$. The respective strength' of the primacy and the recency effect will generally depend on the experimental setup, including the task to be performed. 
Simulations\footnote{\normalsize The code can be found in https://gitlab.com/neuroinf/memorystates/-/tags} are for $2$ arbitrarily (iid) chosen sequences of high-dimensional binary random states of grid size $100 \times 100$ with $q=\frac{1}{3}$ and bundling threshold $\theta = \frac{1}{2}$. Since the size of fluctuations decreases exponentially with dimension, there is no need for sampling over a huge set sequences to obtain statistically significant typical results in high dimensions. 

In Fig. \ref{fig:GenericBehaviour} $\bf (a)$ weights are equal, $\rho=\rho'=1$, the resulting $\cup-$shaped curve is in good agreement with empirical data from Free Recall \cite{murdock1962serial}. On the other hand, in Fig. \ref{fig:GenericBehaviour} $\bf(b)$, weights are different favouring the primacy $\bf R$-state or the recency $\bf L$-state, respectively. This attention asymmetry corresponds to different tasks such as serial recall under forward and backward conditions \cite{oberauer2018benchmarks}.
However, no attempt was made to fit real data. Instead the aim is to show that typical $\cup$-shaped Serial Position Curves are generic outcomes of the model proposed.  

\section{Some concluding remarks concerning the formal states $\bf R$ and $\bf L$}
Given the fact that our model is purely computational in nature, can it still be meaningfully related to neurobiological findings? This particularly concerns its main consequence, which is the existence of the two representations $\bf R$ and $\bf L$.

In light of the fundamental characteristics of high-dimensionality, randomness, and sparseness, we present an elementary algebra that enables the construction of representations of information items. The construction is attributable to interference-like bundling and associative-like binding. In contrast to the typical VSA framework, the bundling operation proposed is not associative. This enables the preservation of serial order during the bundling process. It is thus possible to represent information about items, their contexts, and their serial order. As a consequence of the non-associativity of the bundling operation, a single sequence of items is represented by two states: the $\bf L$-state and the $\bf R$-state. These states are due to left-associative bundling and right-associative bundling, respectively. The two states differ fundamentally in that the  {\bf L}-state can be constructed during an inflow of information on the fly, whereas the  {\bf R}-state can only be constructed after a finite sequence has been presented. The two states implement distinct components of memory: the $\bf L$ state implements the recency component, while the $\bf R$ state implements the primacy component.

These states manifest as discrete yet interrelated entities. This is evidenced by their disparate construction processes on the one hand and their disparate weightings on the other.
While these two states are formal outcomes of our computational model, establishing the primacy and recency components, respectively, one may link both to functional systems such as short-term memory and long-term memory. This ultimately leads to the functioning of corresponding brain structures, including the prefrontal cortex and the hippocampus. These are distinct structures that are nevertheless connected, for example, via the prefrontal-parietal network. It has been demonstrated in experimental studies that short-term memory and long-term memory can be selectively impaired by brain damage. Consequently, impairments of brain structures should result in different effects on the recency and the primary branch. 
The primacy effect should be impaired if the hippocampus is damaged, because this structure is important for the encoding and transfer of information from short-term memory into long-term memory. Under these conditions, the recency effect should remain unaffected. This was shown for rats with dorsal hippocampal lesions \cite{kesner1982serial}, in humans following anterior temporal lobectomy \cite{hermann1996effects}, or in pure amnesic patients \cite{rosenbaum2012amnesias, allen2018classic}. 
The recency effect, on the other hand, is impaired by disruptions in the PFC-parietal network particularly due to damage or dysfunction in the prefrontal cortex, causing slower and less accurate retrieval of recent items, and thus impairs the ability to recall recent events. Deficits in recency effect have been seen in various conditions including neuro-generative disorders such as Alzheimer's disease or frontotemporal dementia causing progressive atrophy in both the PFC and parietal region, leading to significant impairments in working memory and the recency effect; traumatic brain injury, or focal epilepsy, \cite{riley2016role} for review. 
Impairment of the hippocampus may be regarded as disrupting the construction of the $\bf R$ state, finally impacting the primacy branch, while the $\bf L$-remained intact. On the other hand, impairment of the PFC, or PFC-parietal network, affects the construction of the $\bf L$-state and/or the weighting in the decision process.

The main concern of this note is about representing information rather than storing it in some 'memory', from which it can be retrieved upon request. 
In VSA, the retrieval of items from a storage medium is typically achieved through the unbinding process, whereby the item is released from the storage medium. Subsequently, the item is cleaned and read out. In this context, the states $\bf L$ and $\bf R$ would serve as storage media.

In the framework proposed, these states do not function as repositories from which information items are retrieved. They are representations of information and serve as templates against which a query is to be compared. The extent to which a query is represented by a given state serves as the basis for comparison. The aforementioned degree is quantified through the use of mutual information. The mutual information values may be interpreted as representing different degrees of uncertainty. Consequently, the response to whether a specific item is novel to the agent or has been previously presented is contingent upon the decision-making process based on that quantity.

In the context of information gathering, represented by the modification of the existing memory state $\bf M$, the process of bundling could be regarded as a movement in the state space $\mathbb{X}_q$.
With respect to a specific cue, the act of learning could be defined as a change in the relative location of the memory state in relation to that cue. This, in turn, may result in a change in the response of the agent to that cue.
The formal model presented here is consistent with the perspective on "spatial computing" proposed in \cite{lundqvist2023working}. 

Furthermore, the model highlights the significance of concepts such as distributed representation in high dimensions for maintaining the stability of the agent's behaviour. In high dimensions, the sparseness of representations is a crucial factor. Due to the bundling operation proposed, it can be locally controlled, allowing for adjustments concerning capacity and efficiency. In the model outlined, the preservation of sequential information comes at the price that not all items in a long sequence can be recalled with the same degree of accuracy. Alternatively, there is a dynamic 'time frame' within which information items are most prominent. This may be attributed to a reduction in the focus directed towards a specific cue over time, or to the loss of previously encoded information. It could be argued that whilst the agent does not forget, its focus shifts as a result of the acquisition of new information. 

\section{Acknowledgement}
The author is grateful to the anonymous reviewer for valuable criticism that improved the first version of the manuscript. He is also very grateful to Adrian M. Whatley for his help in preparing the manuscript for the publication.

\bibliographystyle{apacite}
\bibliography{NECO-Paper}

\end{document}